\documentclass{article}

\PassOptionsToPackage{numbers, compress}{natbib}
 \usepackage[dblblindworkshop, final]{neurips_2025}
 \usepackage{wrapfig}

\usepackage[utf8]{inputenc} % allow utf-8 input
\usepackage[T1]{fontenc}    % use 8-bit T1 fonts
\usepackage{hyperref}       % hyperlinks
\usepackage{url}            % simple URL typesetting
\usepackage{booktabs}       % professional-quality tables
\usepackage{amsfonts}       % blackboard math symbols
\usepackage{nicefrac}       % compact symbols for 1/2, etc.
\usepackage{microtype}      % microtypography
\usepackage{xcolor}         % colors

\usepackage{amsmath}
\usepackage{amssymb}
\usepackage{amsthm}

\usepackage{graphicx}
\usepackage[ruled,vlined]{algorithm2e}
\usepackage{booktabs}

\usepackage{multirow}

\newcommand{\cut}[1]{}

\usepackage{amsmath,amsfonts,amsthm,bm}

\def\eqref#1{equation~\ref{#1}}
\def\1{\bm{1}}

\def\ry{{\textnormal{y}}}

\def\rvx{{\mathbf{x}}}

\def\rvz{{\mathbf{z}}}

\def\vw{{\bm{w}}}
\def\vx{{\bm{x}}}
\def\vy{{\bm{y}}}
\def\vz{{\bm{z}}}

\DeclareMathAlphabet{\mathsfit}{\encodingdefault}{\sfdefault}{m}{sl}
\SetMathAlphabet{\mathsfit}{bold}{\encodingdefault}{\sfdefault}{bx}{n}

\def\gD{{\mathcal{D}}}

\def\gF{{\mathcal{F}}}

\def\gH{{\mathcal{H}}}

\def\gL{{\mathcal{L}}}

\def\gX{{\mathcal{X}}}
\def\gY{{\mathcal{Y}}}

\DeclareMathOperator*{\argmax}{arg\,max}
\DeclareMathOperator*{\argmin}{arg\,min}

\title{Computing Strategic Responses to Non-Linear Classifiers}
\workshoptitle{Unifying Perspectives on Learning Biases}
\author{%
  Jack Geary\\
  School of Informatics\\
  University of Edinburgh\\
  Edinburgh, United Kingdom\\
  \texttt{jack.geary@ed.ac.uk} 
  \And
  Boyan Gao\\
  Department of Engineering Science\\ 
  University of Oxford\\
  Oxford, United Kingdom\\
  \texttt{boyan.gao@eng.ox.ac.uk} 
  \And
  Henry Gouk\\
  School of Informatics\\
  University of Edinburgh\\
  Edinburgh, United Kingdom\\
  \texttt{henry.gouk@ed.ac.uk}
}

\begin{document}

\maketitle

\newcommand{\keypoint}[1]{\textbf{#1}\qquad}

\begin{abstract}
% For the purposes of this workshop, let's try to appeal to their goals
% Strategic Classification models the interaction between a \textit{Learner} that learns a classifier, and \textit{Agents} who aim to be positively classified. Agents are empowered to manipulate their state to achieve their objective, with the optimal feasible manipulation called the \textit{best response}.
We consider the problem of strategic classification, where the act of deploying a classifier leads to strategic behaviour that induces a distribution shift on subsequent observations. Current approaches to learning classifiers in strategic settings are focused primarily on the linear setting, but in many cases non-linear classifiers are more suitable. A central limitation to progress for non-linear classifiers arises from the inability to compute best responses in these settings. We present a novel method for computing the best response by optimising the Lagrangian dual of the Agents' objective. We demonstrate that our method reproduces best responses in linear settings, identifying key weaknesses in existing approaches. We present further results demonstrating our method can be straight-forwardly applied to non-linear classifier settings, where it is useful for both evaluation and training.
%In the final version should probably point to the code at the end of the abstract
\end{abstract}

\section{Introduction and Related Work}
% Core argument of the paper:
% - In realistic scenarios involving classification, a non-linear classifier is often preferable to a linear classifier (in terms of accuracy).
% - Current approaches to Strategic Classification only have solution guarantees for simple (linear) classifiers. 
% - A limit to current iterative approaches is that standard gradient-based optimisation methods only compute \partial_L/\partial_\theta. So the current solution approaches, such as the iterative approach proposed in Perdomo, only align with the true gradient solution in limited settings (explain this?)
% - An approach from other fields that address this issue is Implicit Gradients. This formula allows you to more accurately account for the Learner's ability to the respond to the best response in the update calculation. 
% Implicit Gradients does not make any restrictive assumptions about the family of models to which it can be applied. As such learning in this setting frees the restriction that typically results from following other learning approaches.
% We propose a new learning procedure, Strategic Training, that incorporates Implicit Gradients in the Strategic Classifier Learning procedure. 
%
Biased training data can have a significant adverse impact on the accuracy and fairness of models when they are applied to future observations that come from a different distribution. In classifier learning, one possible source of bias arises from tactical misrepresentation by people who interact with the classifier; those with a preference for a certain classification, and a knowledge of the classifier that is being used, have an incentive to misrepresent their state to receive a favourable outcome. This dynamic can arise in many sociotechnical situations, spanning from universities deciding which students to enrol as far as institutions determining how to distribute aid and resources in times of need.

The Strategic Classification \citep{hardt2016, bruckner2011} literature examines the setting where a \textit{Learner} is attempting to classify a population of \textit{Agents} in that can independently perturb their representations in response to the learned classifier. The principal objective of the Learner is to develop a classifier that is robust to the Agents' perturbations, thereby reducing their ability to undermine the overarching goal of the Learner. This is typically formulated in the language of game theory: the Learner takes the role of a leader in a Stackelberg game, and the Agents follow with their best response. Computing the best response presents a considerable challenge in Strategic Classification; its definition is typically discontinuous and non-differentiable in all but the simplest cases, which makes it unwieldy to work with. As a consequence of this impediment, most of the research that has been done in Strategic Classification has been focused on linear classifiers, where one has a closed form expression for the best response \citep{levanon2021, levanon2022, eilat2022}. This limits the diversity and the scope of the research that can be performed, as well as the degree to which developments in this area can be plausibly adopted and applied in real-world settings.

One algorithm that can be used for approximating the best response to non-linear classifiers is Repeated Empirical Gradient Descent (REGD), as proposed in \citet{perdomo2020}. Arising from the related field of Performative Prediction, the authors demonstrate that it can be applied to the Strategic Classification setting. While the method in \citet{perdomo2020} can be used for non-linear classifiers, they focus on the case of linear classifiers. Subsequent works, such as \citet{izzo2021, mofakhami2023, zhong2025}, have iterated on this algorithm to work on a more generalised performative settings. However, there approaches utilise methods such as counterfactual reasoning to make inferences about the behaviour of the best response, and generally avoid explicitly computing it.

In this work we present a method for computing strategic responses to non-linear classifiers that is less conservative and more cost-efficient than the approach of \citet{perdomo2020}. Building on recent iterative methods from the field of Performative Prediction, we also demonstrate that our method results in a learning algorithm that is capable of producing strategically robust classifiers for a range of differentiable classifier families.

\section{Strategic Classification and the Best Response}
% Why should this work? Lay out the Strategic Classification problem and typical formulation
% Highlight directly applying partial gradient computation on the best response isn't accurate (but it typically what arises from iterative approaches). Lay out Implicit Gradient. Highlight that the Implicit Gradient is the common approach to deal with this in Adversarial context.
For a feature space, $\gX$, and label space, $\gY = \{-1,1\}$, let $\gD$ be a probability distribution on their Cartesian Product, $\gX \times \gY$. Given a class of models, $\gH \subset \mathbb{R}^\gX $, the usual goal when training a classifier is to find some model, $f \in \gF = \{\vx \mapsto \operatorname{sgn}(h(\vx)) : h \in \gH\}$, that achieves high accuracy. Given access to a training dataset, $S = \{(\rvx_{i}, \ry_{i})\}_{i=1}^{n}$, consisting of independent and identically distributed samples from $\mathcal{D}$, this goal is often addressed by finding a model that minimises the empirical risk,
\begin{equation}
    \hat{f} = \argmin_{f \in \mathcal{F}} \dfrac{1}{n} \sum_{i=1}^{n}l(f(\rvx_{i}), \ry_{i}),
\end{equation}
where $l$ would ideally be the zero--one loss, but is more typically chosen to be the hinge loss in order to facilitate tractable optimisation.

In many realistic classification scenarios $\mathcal{D}$ is not representative of the data that a trained classifier will actually be deployed on \citep{quinonero2022dataset,rosenfeld2023a}. Distribution shift can adversely impact the accuracy of $f$, since $f$ will not necessarily be optimal under the new data distribution. In a strategic setting, distribution shift can arise because Agents corresponding to some data point, $(\vx, y)$, may be motivated to manipulate their representations, $\vx$, in order to obtain a positive classification from $f$, irrespective of their true label, $y$. In the strategic classification literature this is typically modelled using an idealised best response function \citep{hardt2016},
\begin{equation}
    \Delta_f(\vx) = \argmax_{\vz \in \gX} f(\vz) - c(\vx, \vz).
    \label{eqn:best_response}
\end{equation}
Crucially, it is assumed that an Agent must pay some cost, $c(\vx, \vz)$, to manipulate their representation from $\vx$ to $\vz$. This cost is assumed to satisfy several natural properties: no manipulation should result in no cost, and the cost should be subadditive, $c(\vx,\vz) \leq c(\vx,\vy) + c(\vy,\vz)$. Various extensions of this standard setup exist that allow for, e.g., different levels of information to be revealed to the Agents \citep{ghalme2021, cohen2024} or for Agents to not necessarily favour positive classification \citep{levanon2022}.

This strategic interaction between the Agents and the Learner results in a breakdown in performance of the classifier \citep{levanon2021}. To address this, algorithms have been designed to support training of classifiers that are robust to such misrepresentations \citep{levanon2021, sundaram2023, perdomo2020}. In particular, the idea of adapting the empirical risk to include this idealised best response is natural. This has lead to the Strategic Empirical Risk Minimisation (SERM) approach \citep{sundaram2023},
\begin{align}
    \label{eqn:strategic_classification}
    \hat{f} = \argmin_{f \in \mathcal{F}} & \quad \dfrac{1}{n} \sum_{i=1}^{n}l(f(\Delta_{f}(\rvx_{i})), \ry_{i}) \\
    \operatorname{subject\,to}& \quad \Delta_{f}(\rvx_i) = \argmax_{\vz \in \mathcal{X}} f(\vz) - c(\rvx_i, \vz), \nonumber
\end{align}
a bilevel optimisation problem, where the $\argmin$ over $f$ is referred to as the upper level and the $\argmax$ over $\vz$ as the lower level. For a linear model parameterised by $\vw$ and the Euclidean distance cost function, this collapses to a single level problem due to the closed form best response \citep{dong2018, perdomo2020},
\begin{equation}
    \label{eqn:linear-exact}
    \Delta_{f}(\vx) = \begin{cases}
    \vx - \vw \frac{h(\vx)}{\|\vw\|_2} & \text{if $-2 \leq h(\vx) < 0$}\\
    \vx & \text{otherwise}.
    \end{cases}
\end{equation}
In cases where a closed form is not available, one can employ the Repeated Empirical Gradient Descent (REGD) approach of \citet{perdomo2020} to obtain an approximation of the empirical risk minimiser, $\hat{f}$. This requires black box access to a routine that computes a strategic response to some classifier, $f$.

\section{Responding to Non-Linear Classifiers}
\label{sec:lagrangian_best_response}
% Intro to Lagrangians, KKT and how to rewrite the best response in this format
When the underlying classifier is non-linear, or the cost is less amenable to optimisation than the squared Euclidean distance, we cannot expect to find a global optimum of Equation \ref{eqn:best_response}. In this Section we discuss methods for computing acceptable approximations to the best response.

\keypoint{Gradient Descent Response}
% Strategic Classification requires solving a constrained optimisation problem for the inner-objective. 3 constraints; maximise utility, minimise cost, cost must be less than 2
% A popular approach to solving optimisation problems like best response is closed form or differentiation. The issue with this inner objective is that a binary classifier f is not differntiable.
Gradient-based methods are a common approach for solving optimisation problems such as Equation \ref{eqn:strategic_classification}. Such methods involve using gradient ascent to solve the lower level maximisation, and gradient descent to solve the upper level minimisation. \citet{perdomo2020}, propose an iterative gradient-based method that they show can be used to approximate solutions to Strategic Classification problems.\footnote{They prove that the convergence of their method to optimal solutions is limited to a narrow family of classifiers} However, we note that, since $f$ is a binary classifier, it is not differentiable, and so gradient methods cannot be directly applied to the problem as presented in Equation \ref{eqn:strategic_classification}.

% You could replace f with f' (continuous), but then you lose one of the constraints. Also since f' continuous, there is now a relationship between utility and cost, so the magnitude of f may supress cost (this comes up in figure later)
In the case where $\gH$ is a class of differentiable models, one approach to allow for gradient-based solutions concepts is to replace $f$ in the lower objective with the corresponding differentiable $h$,
\begin{equation}
    \Delta_f^{GD}(\vx) = \argmax_{\vz \in \gX} h(\vz) - c(\vx, \vz).
    \label{eqn:gradient_best_response}
\end{equation}
This is effectively what is done in \cite{perdomo2020}. However, as a consequence of this relaxation, one of the constraints on the best response is relaxed; in the best response of Equation \ref{eqn:best_response}, solutions trade off maximising $f(\vz)$, minimising the cost $c(\vx, \vz)$, and constraining the cost to be less than two ($c(\vx, \vz)\leq2$). This last constraint arises from the interaction between the utility and the cost terms in the optimisation; $f$ is binary and the maximum the utility can be improved by is two. If the cost to realise this exceeds two then the Agent would realise a higher objective by letting $\vz = \vx$. This constraint does not necessarily apply in the case where $f$ is replaced with $h$.
%In the case when $f$ is linear and the cost is the Euclidean distance, this problem can be circumvented as the distance of a point from the classifier decision boundary has a closed-form solution, and so it is possible to determine if $\vx$ is close enough to the decision boundary to game it. But in general this is not the case.

\keypoint{Lagrangian Dual Response}
% The problem can be reformulated as a bilevel optimisation problem. This eliminates the interaction between utility and cost.
In this work we consider an alternative formulation of the lower level objective. By treating it explicitly as a constrained optimisation problem, where the objective is to minimise the cost, we can write Equation \ref{eqn:best_response} equivalently as,
\begin{align}
        \label{eqn:best_response_bilevel_form}
        \Delta_{f}(\rvx) = \argmin_{\rvz \in \mathcal{X}} & \quad c(\rvx, \rvz),\\
        \operatorname{subject\,to} &\quad h(\rvz) \geq 0,\nonumber \\
        & \quad c(\rvx, \rvz) \leq 2. \nonumber
\end{align}
This formulation captures the same objectives that applied in Equation \ref{eqn:best_response}, but it does so without depending on the interaction between the utility and the cost during the optimisation. In practice, one must constrain $h(\vz) \geq \epsilon$ for some small $\epsilon$ to ensure the inequality holds strictly.

% Problems in this context can be solved using KKT methods. These involve introducing lagrange multipliers. This gives us the resulting optimisation problem that is amenable to being solved by differential methods.
The problem formulation in Equation \ref{eqn:best_response_bilevel_form} is amenable to Karush-Kuhn-Tucker (KKT) solution methods \citep{kuhn2016, gordon2012}. Under this formulation we can replace the objective in Equation \ref{eqn:best_response_bilevel_form} with the Lagrangian dual,
\begin{equation}
    \label{eqn:lagrangian}
    \gL(\vx, \vz, \lambda, \mu) = c(\vx, \vz) - \lambda(h(\vz) - \epsilon) + \mu(c(\vx, \vz) - 2),
\end{equation}
where $\lambda, \mu \in \mathbb{R}$ are Lagrange multipliers. This is used to define an optimisation problem,
\begin{align}
    \label{eqn:lagrangian_best_response}
    \Delta_{f}^{LD}(\rvx) = \argmin_{\rvz} \max_{\lambda, \mu} & \quad \mathcal{L}(\rvx, \rvz, \lambda, \mu)\\
    \operatorname{subject\,to} &\quad \lambda, \mu \geq 0, \nonumber
\end{align}
that can be solved with projected gradient ascent-descent. The projection function simply clamps the $\lambda$ and $\mu$ Lagrange multipliers to ensure they remain non-negative. The solution to the Langrangian dual optimisation converges to a locally optimal solution whilst enforcing the constraints. In the case where all local optima are global optima---e.g., if $h$ and $c$ are convex in $\vz$---this approach is guaranteed to compute the best response.

\keypoint{Post-Response Checks}
\label{sec:post_response_checks}
Valid responses must satisfy two invariants, $f(\Delta_{f}(\rvx))>0$ and $c(\rvx, \rvz)<2$. Both approaches described above may violate these invariants in some instances. The Gradient Descent response will sometimes incur more cost than is rational, given the reward gained from obtaining a positive classification. The Lagrangian Dual may violate one of these constraints if there is no feasible point. To avoid these issues, we manually check that both invariants are satisfied and return the original $\vx$ if they are not.

\section{Experimental Results}
% \subsection{Implementation Details}
% % Introduce the settings for the experiments
% In Section \ref{sec:lagrangian_best_response} we have presented a method for reformulating the classical best response definition (Equation \ref{eqn:best_response}) such that it is amenable to gradient-based solutions. In order to apply the REGD algorithm to the problem we also require the use of a differentiable loss function for Equation \ref{eqn:strategic_classification} (since $\ind$ is not differentiable). In this work we use the hinge loss as a surrogate, as proposed in \cite{levanon2022}. Therefore Equation \ref{eqn:strategic_classification} becomes:
% \begin{equation}
%     \begin{split}
%         \hat{f} &:= \argmin_{f \in \mathcal{F}} \dfrac{1}{n} \sum_{i=1}^{n}\max(0, 1-f(\Delta_{f}(\rvx_{i})), \ry_{i}) \\
%         &s.t. \quad \Delta_{f}(\rvx) = \argmin_{\rvz} \max_{\lambda, \mu} \mathcal{L}(\rvx, \rvz, \lambda, \mu)
%     \end{split}
%     \label{eqn:strategic_classification_applied}
% \end{equation}

% Not sure if we really need to include learning rates or something here.

% Could probbaly throw in an algorithm block in here explaining what the REGD algorithm is.

%% !!TODO!!: I think this NEEDS to be included
% Should also highlight the post-optimisation checks (so we only keep points that satisfy the criteria of f(z)>f(x) and c(x,z)<2

\keypoint{Responding to Linear Models}
% Demonstrate that the Lagrangian Method can accurately compute best response in scenario with known "correct" answer
In order to qualitatively examine and compare different response computations methods we examine their impact on a simple toy dataset. We consider a two dimensional dataset comprised of two Gaussians; one corresponding to the negative class, and the other to the positive. Both classes have the same number of data points. We train a linear support vector machine (SVM) on the dataset, without taking any effort to make the classifier strategically robust. We consider three methods for computing strategic responses: the exact solution given in Equation \ref{eqn:linear-exact}, the Gradient Descent response in Equation \ref{eqn:gradient_best_response}, and the Lagrangian Dual response from Equation \ref{eqn:lagrangian_best_response}. For all three approaches we use the Euclidean distance as the cost function. 

\begin{figure*}[t]
    \centering
    \includegraphics[width=0.235\linewidth]{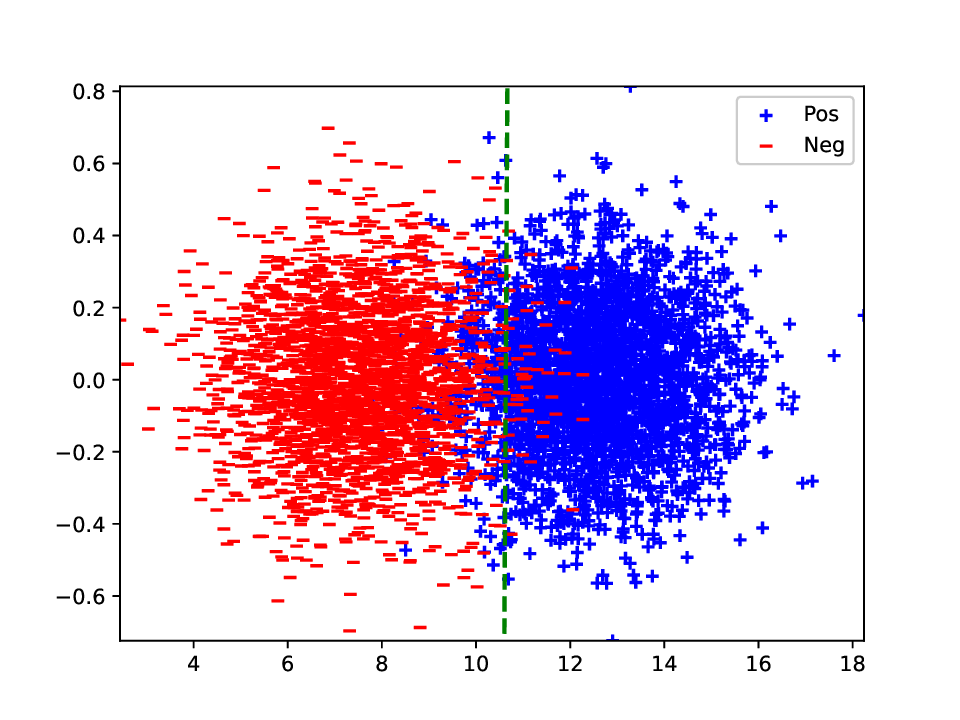}
    \includegraphics[width=0.235\linewidth]{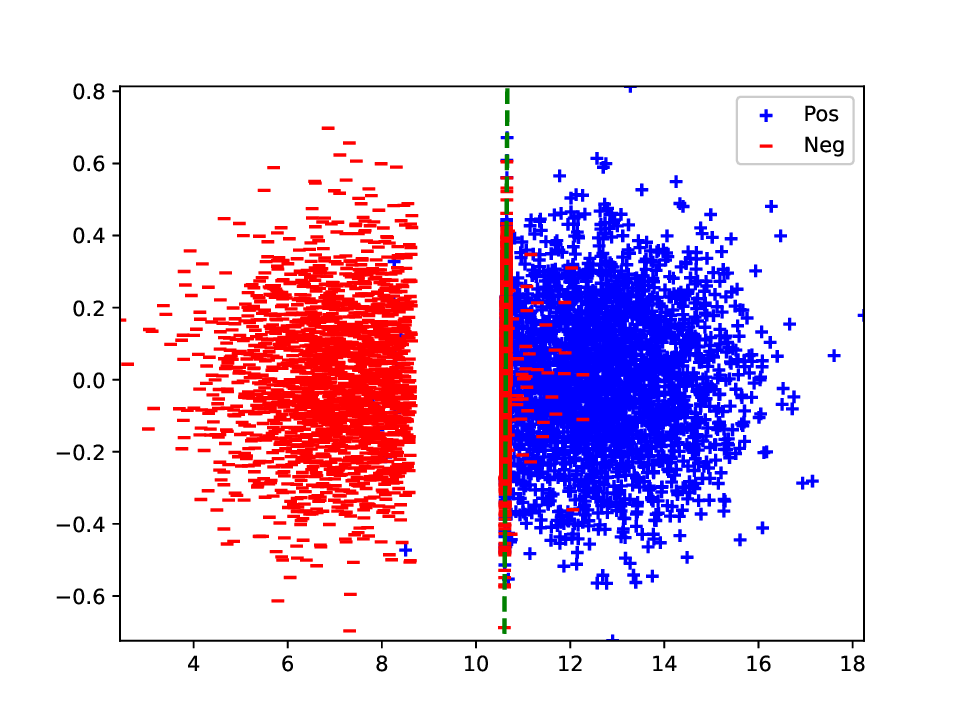}
    \includegraphics[width=0.235\linewidth]{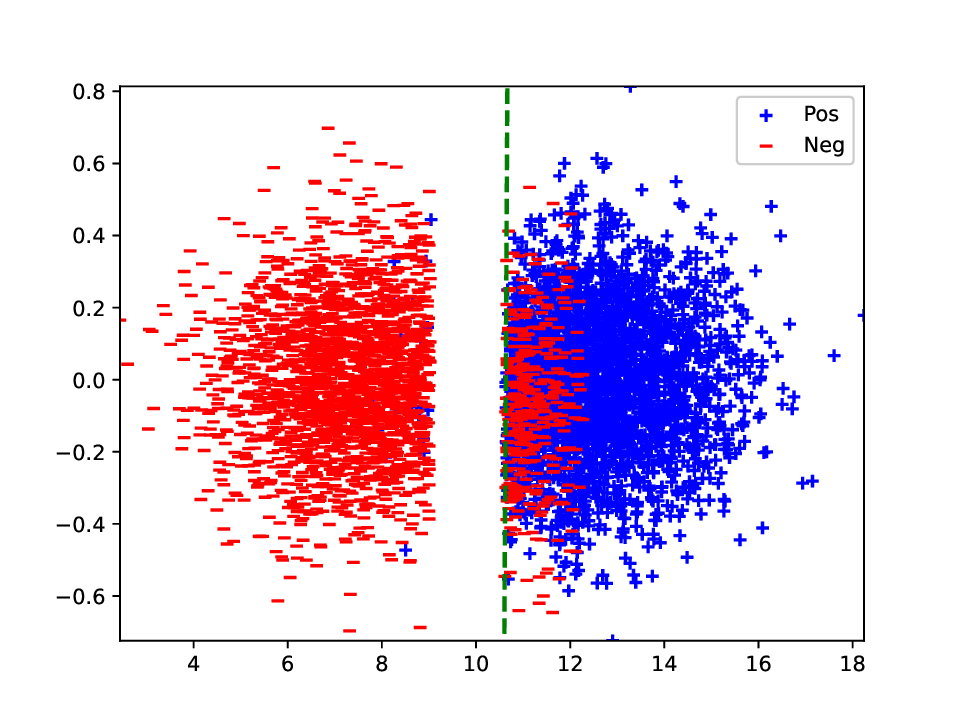}
    \includegraphics[width=0.235\linewidth]{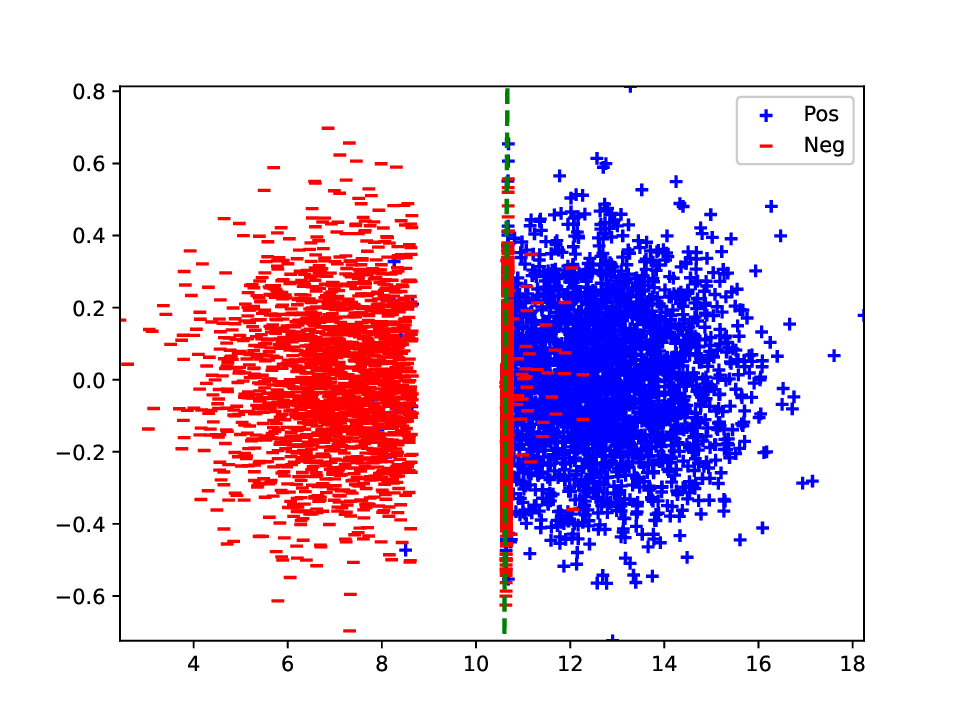}
    \caption{Various strategic responses to a fixed linear classifier applied to Gaussian dataset; Left: Linear SVM decision boundary; Middle Left: Ground Truth Response; Middle Right: Gradient Descent Response; Right: Lagrangian Dual Response (Ours)}
    \label{fig:exp1_response_on_ball}
\end{figure*}

Figure \ref{fig:exp1_response_on_ball} provides a visual comparison of these response methods on the Gaussians dataset. The left pane shows the unperturbed dataset along with the location of the SVM decision boundary. The Left-Middle pane demonstrates the best response behaviour arising from using the closed form expression in Equation \ref{eqn:linear-exact}. The Middle-Right and the Right panes are the Gradient Descent and Lagrangian Dual responses, respectively. Comparing with the optimal response, we observe that the Gradient Descent approach is susceptible to two types of errors: (i) there are some points that move too far over the decision boundary, meaning they incur more cost than is necessary; and (ii) some points that could pay a cost close to, but still less than, two are not manipulated. In contrast to this, our Lagrangian Dual approach is able to exactly replicate the behaviour of the true best response.

% We compare this result with the result from computing the response to $f$ using a gradient-based to solve Equation \ref{eqn:best_response} with $u(\rvz, f) = \tilde{f}(\rvz) \in \mathbb{R}$ (where $f(\rvz) = \text{sign}(\tilde{f}(\rvz))$). In this setting Equation \ref{eqn:best_response} has a closed-form solution at $\Delta_{f}(\rvx) = \rvx + \epsilon w$. The amount that points are moved according to this definition depends on the magnitude of the model weights. As such it can fail to identify some points which could feasibly game the classifier (if $||w||<2$). Similarly, if $||w||>2$, then this response will find no points that can game the classifier with the constraint that $c(\rvx, \rvz)<2$. Figure \ref{fig:exp1_response_on_ball} (Middle Right) shows the result of applying this response definition to our toy example; we see the response has underestimated the set of points that can game the classifier, and the points that have gamed the classifier have overshot the decision boundary, so it has also failed to minimise the cost.

% Finally we consider the results of computing the response by optimising the Lagrangian proposed in Equation \ref{eqn:lagrangian_best_response}. This definition explicitly aims to find the minimum cost solution, while getting as many points as possible over the decision boundary. As a consequence of these joint objectives, Figure \ref{fig:exp1_response_on_ball} (Right) shows that the response very well approximates the ground-truth solution. 

\begin{figure*}[t]
    \centering
    \includegraphics[width=0.235\linewidth]{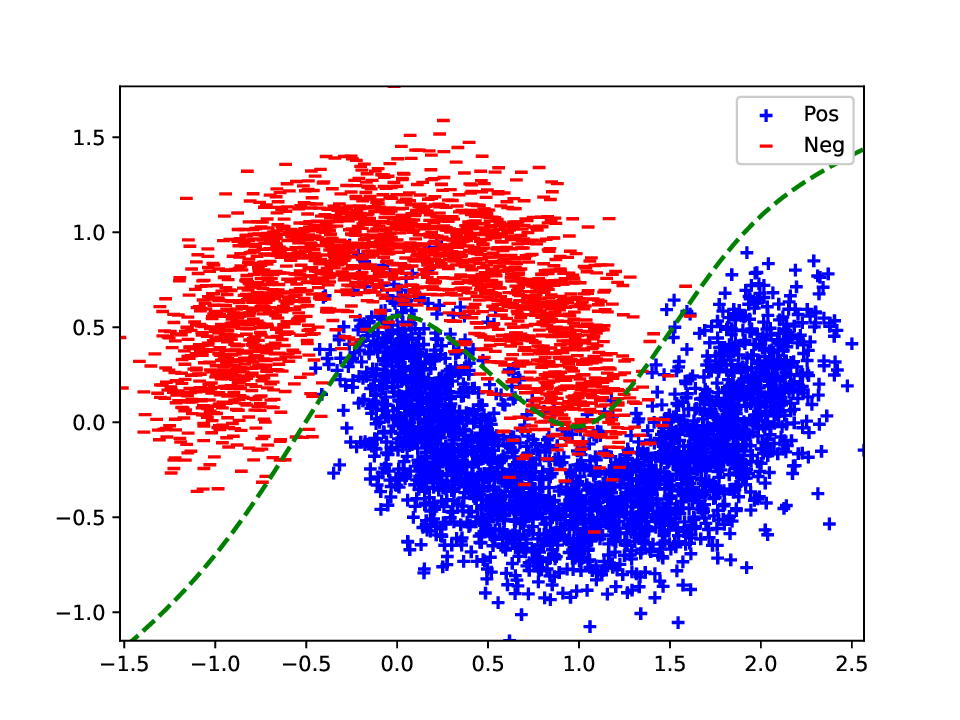}
    \includegraphics[width=0.235\linewidth]{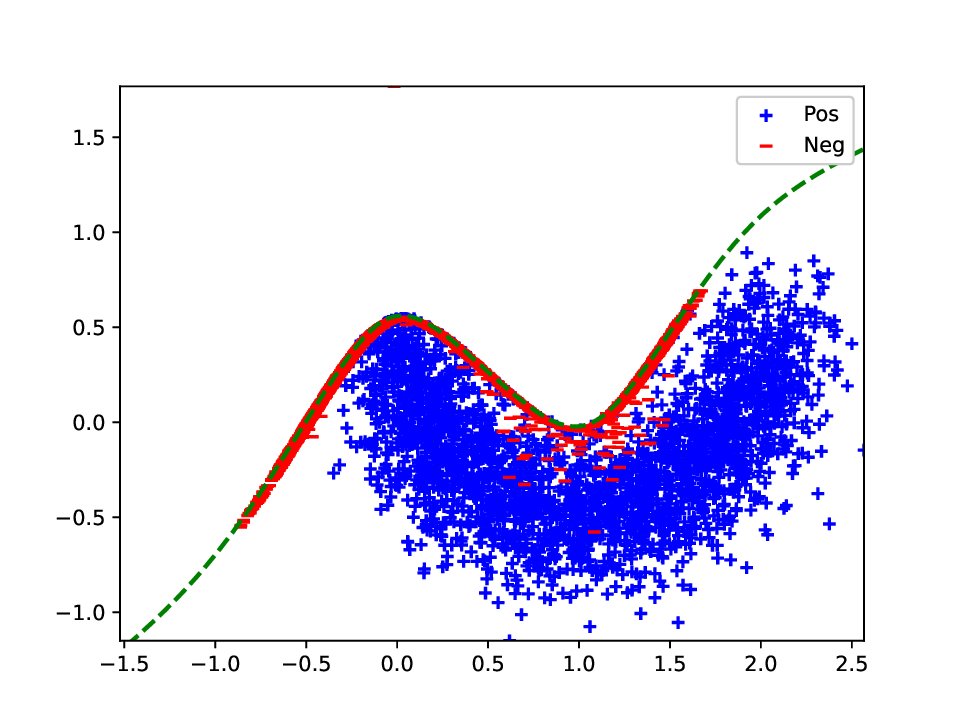}
    \includegraphics[width=0.235\linewidth]{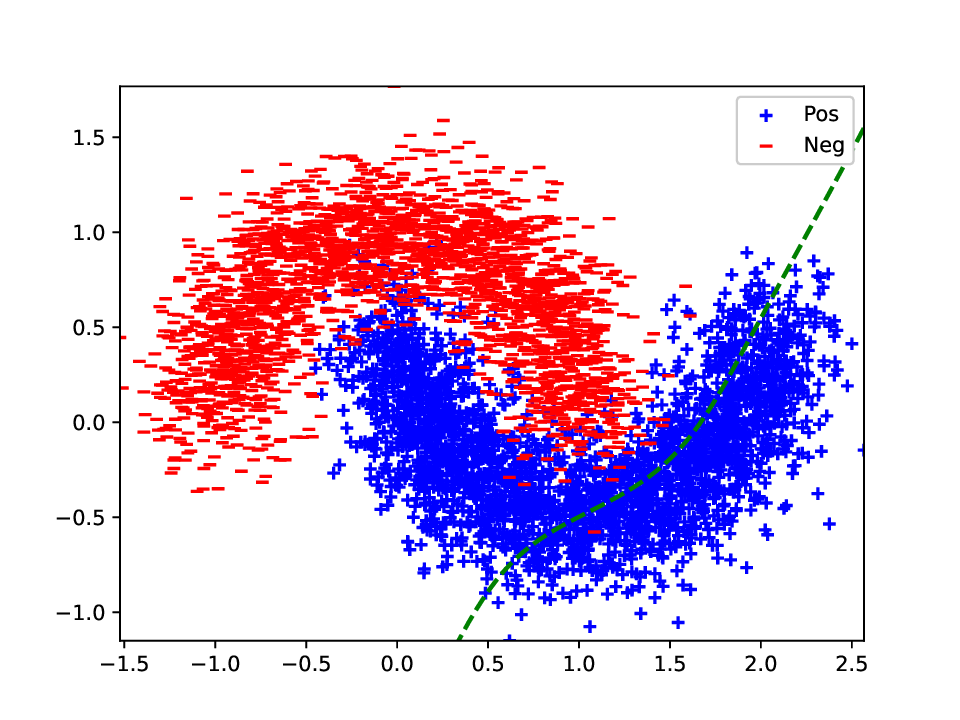}
    \includegraphics[width=0.235\linewidth]{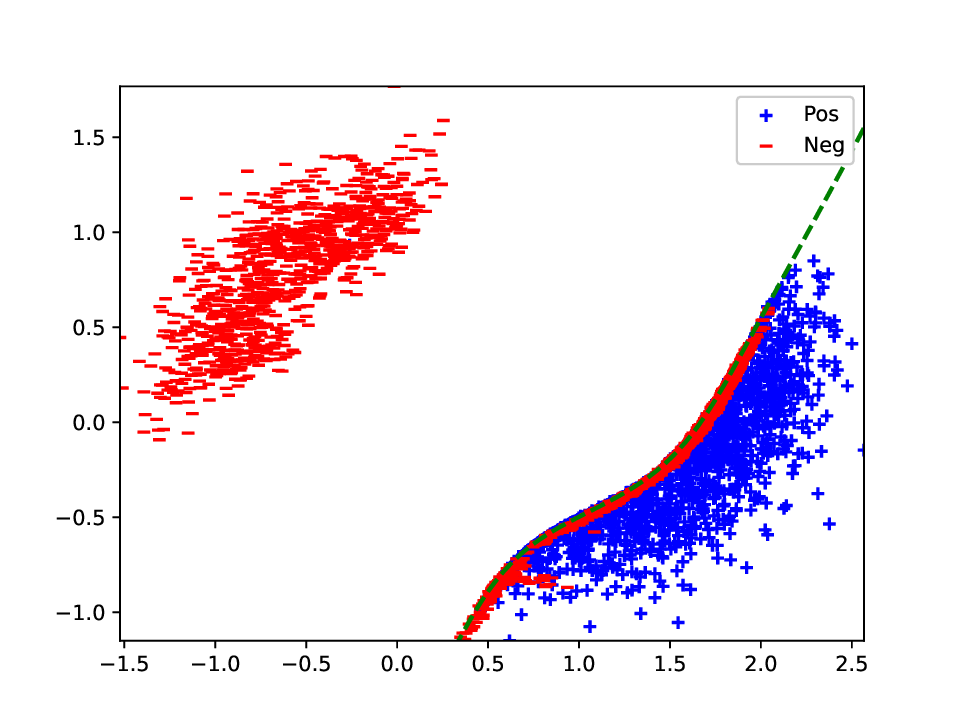}
    \caption{Left: MLP Classifier trained on twin moons with ERM; Left Middle: Lagrangian Response to ERM MLP model; Middle Right: MLP Classifier trained with REGD method; Right: Lagrangian Response to REGD MLP model.}
    \label{fig:nonlinear_model_response_on_twin_moons}
\end{figure*}

\keypoint{Responding to Non-Linear Models}
% Quadratic Classifier on Toy Dataset/MLP on Twin Moons
% Simple demo showing that our method does allow us to train non-linear classifiers.
% Want to find the simplest example where there doesn't exist an obvious closed-form definition for the best response, and the context violates the iterative solution approach and it is possible to easily (preferably visually) assess the quality of the classifiers => Perdomo method is fallible.
% Toy dataset like ring dataset with quadratic classifier, or something more classical like twin moons with MLP might look impactful.
%Prior works have presented methods for training strategically robust classifier models (\cite{perdomo2020, levanon2022}). In particular, \cite{perdomo2020} propose REGD as an iterative method for training a model by computing the gradient of a differentiable loss with respect to $\Delta{f}$. However, in the case of nonlinear models, the best response does not have a closed-form solution. So it has not been possible to use REGD to produce strategically robust nonlinear models.
We investigate qualitatively whether our Lagrangian Dual response is able to successfully respond to a classifier trained with ERM, and another trained using REGD with our approach also used as the response function. This is achieved by training a small Multi-Layer Perceptron (MLP) model on the twin moons toy dataset.

Figure \ref{fig:nonlinear_model_response_on_twin_moons}  shows the results of this experiment. The Left pane visualises the ERM decision boundary and the unperturbed dataset. While this is optimal in absence of strategic behaviour, all of the negative points are close to the decision boundary, and our strategic response is therefore able to move them to lie on the positive side of the decision boundary, as demonstrated in the Middle Left pane.  The Middle Right pane depicts the decision boundary when the model is trained with REGD. Accuracy on the unperturbed data points is poor, but when the strategic behaviour is applied in the Right pane, we see that all the positive points are classified correctly but only a subset of the negatively labelled points successfully manipulate their features.

\begin{table}[t]
\centering
\parbox{0.545\linewidth}{\caption{Percentage of correct predictions (mean $\pm$ standard error) made by linear models (top), ICNN model (middle), and MLP model (bottom) on the Give Me Some Credit Dataset when different strategic response methods are used during training (rows) and testing (columns).}
\label{table:give_me_some_credit_accuracy}
\resizebox{\linewidth}{!}{\begin{tabular}{c|ccccc}
\toprule
 & $\Delta_f$ & Identity & Gradient & Lagrange \\
\midrule
 \multirow{3}{*}{Linear} & I & $59.00 \pm 0.38$ & $52.06 \pm 0.39$ & $50.34 \pm 0.39$ \\
 & GD & $47.46 \pm 0.39$ & $49.55 \pm 0.39$ & $49.84 \pm 0.39$ \\
 & LD & $57.65 \pm 0.38$ & $57.81 \pm 0.38$ & $50.15 \pm 0.39 $ \\
\midrule
 \multirow{3}{*}{ICNN} & I & $63.33 \pm 0.37$ & $67.62 \pm 0.36$ & $50.04 \pm 0.39$ \\
 & GD & $63.73  \pm 0.37$ & $63.78 \pm 0.37$ & $50.86 \pm 0.39$ \\
 & LD & $64.65 \pm 0.37$ & $63.97 \pm 0.37$ & $51.21 \pm 0.39$\\
 \midrule
\multirow{3}{*}{MLP}& I & $73.59 \pm 0.34$ & $50.41 \pm 0.39$ & $50.00 \pm 0.39$ \\
 & GD & $69.16 \pm 0.36$ & $50.21 \pm 0.39$ & $50.08 \pm 0.39$\\
 & LD & $62.49 \pm 0.37$ & $62.64 \pm 0.37$ & $54.33 \pm 0.39$\\
\bottomrule
\end{tabular}}}
\quad
\parbox{0.41\linewidth}{
\caption{Percent of points gamed (mean $\pm$ standard error) for linear models (top), ICNN model (middle), and MLP model (bottom) on the Give Me Some Credit Dataset for different strategic response methods.}
\label{table:give_me_some_credit_gaming}
\resizebox{\linewidth}{!}{\begin{tabular}{c|ccccc}
\toprule
 & $\Delta_f$ & Gradient & Lagrange \\
\midrule
 \multirow{3}{*}{Linear} & I & $42.59 \pm 0.38$ & $48.72 \pm 0.39$ \\
 & GD & $60.74 \pm 0.38$ & $66.00 \pm 0.37$ \\
 & LD & $75.32 \pm 0.33 $ & $88.84 \pm 0.24$ \\
\midrule
 \multirow{3}{*}{ICNN} & I & $26.60 \pm 0.34$ & $79.80 \pm 0.31$ \\
 & GD & $22.47 \pm 0.32$ & $57.91 \pm 0.38$ \\
 & LD & $16.99 \pm 0.29$ & $58.43 \pm 0.38$ \\
 \midrule
 \multirow{3}{*}{MLP}& I & $60.76 \pm 0.38$ & $61.16 \pm 0.38$ \\
 & GD & $61.90 \pm 0.38$ & $62.16 \pm 0.38$ \\
 & LD & $44.18 \pm 0.38$ & $69.58 \pm 0.36$ \\
\bottomrule
\end{tabular}}}
\end{table}

\keypoint{Quantitative Comparison}
%One area where classifiers can be vulnerable to gaming behaviour is finance; banks rely on credit scoring models to evaluate loan applicants and determine who gets approved for loans. As such loan applicants have an incentive to misrepresent their state to the bank in order to appear eligible (a classical example of this behaviour is people opening multiple credit cards to inflate their credit score) \citep{hurley2016}. In these settings it is important to be able to train classifiers that are strategically robust.
We provide a quantitative comparison on the GiveMeSomeCredit dataset \citep{GiveMeSomeCredit}, a popular testbed in the Strategic Classification literature due to the incentives and capabilities of loan applicants to strategically manipulate their features \citep{hurley2016}. The dataset details information about people with the objective of predicting whether they will experience imminent financial distress. %\citet{perdomo2020} previously used this dataset to train and evaluate strategically robust linear models \citep{perdomo2020}.\footnote{We note that in \citep{perdomo2020} the authors train their model under the assumption that only a subset of the features are vulnerable to strategic behaviour. However, for our experiments we consider all dataset features as being strategic.} 
Three types of classifiers are trained: linear, MLP, and Input Convex Neural Networks \citep{amos2017}. ICNNs are MLPs that are constrained to ensure they implement functions that are convex with respect to the input features. We train the models to be strategically robust by minimising the cross entropy using the REGD training algorithm. We instantiate REGD with three different response functions: the identity, $\Delta_{f}^{I}(\vx) = \vx$; the Gradient Descent response $\Delta_{f}^{GD}$; and the Lagrangian Dual response $\Delta_{f}^{LD}$. Each model is evaluated by examining how robust they are to the each of the response definitions.

Table \ref{table:give_me_some_credit_accuracy} contains the results of our experiments. Each row of the table corresponds to a different model instance, and each column refers to the response definition used in the evaluation. We find that robustness promoting training can adversely impact on the model performance on the unperturbed dataset. However, models trained to be robust to the Lagrangian Dual response are consistently more strategically robust than models trained under either Identity or Gradient Descent responses.
%Comparing the Gradient and Lagrange columns of Table \ref{table:give_me_some_credit_accuracy} we see consistently that the Lagrangian-based response is better able to game the classifiers. This demonstrates that it better approximates the ``true'' best response behaviour than the gradient-based approach. 
This result is support by Table \ref{table:give_me_some_credit_gaming}, which shows the proportion of the dataset that successfully responds the classifier in each experiment. We see that, for all model types, the Lagrangian Dual response consistently identifies more points that can respond the classifier than the Gradient Descent approach. The only points that can be perturbed by either response method are points which would be perturbed under the ``true'' best response, $\Delta_{f}$. This is a consequence of the post-response checks described in Section \ref{sec:post_response_checks}. This indicates that the Lagrangian Dual response approximates the best response with greater recall than the Gradient Descent response.

%Takeaway: Our method does successfully train models that are robust to strategic behaviour

\section{Discussion \& Conclusion}
Strategic behaviour is a very accessible means of inducing bias in data distributions, which can in turn undermine the accuracy and fairness of classifiers applied to this data. If it is not appropriately addressed it can facilitate malicious actors in compromising the ability of institutions to appropriately allocate resources. Therefore it is important that methods exist for producing classifiers that are robust to these behaviours. In this work we have presented a novel method for approximating strategic responses to non-linear classifiers, and have demonstrated that it can be effectively utilised to produce strategically robust non-linear classifiers.

Our intention in this work is to highlight the benefits that can be realised by producing models that are robust to strategic behaviour. However, as is apparent from our results (Figure \ref{fig:nonlinear_model_response_on_twin_moons} and Table \ref{table:give_me_some_credit_accuracy}), producing models that are robust to strategic behaviour can have the inadvertent consequence of negatively misclassifying some people. These people, who otherwise may not have been inclined to behave strategically, thereby have no option but to attempt to manipulate the classifier, or else risk misclassification. This can put undue financial (or otherwise) burden on people. While some work has been done to explore this consequence in strategic settings (e.g., \citep{milli2019}), an adequate resolution has yet to be found.

\begin{ack}
This work was funded by NatWest Group via the Centre for Purpose-Driven Innovation in Banking. This project was supported by the Royal Academy of Engineering under the Research Fellowship programme.
\end{ack}

\clearpage

\bibliographystyle{abbrvnat}
\bibliography{bibliography}

\end{document}